\documentclass{article}

\usepackage{PRIMEarxiv}
\usepackage{amsmath}
\usepackage[utf8]{inputenc} 
\usepackage[T1]{fontenc}    
\usepackage{hyperref}       
\usepackage{url}            
\usepackage{booktabs}       
\usepackage{amsfonts}       
\usepackage{nicefrac}       
\usepackage{microtype}      
\usepackage{lipsum}
\usepackage{fancyhdr}       
\usepackage{graphicx}       
\graphicspath{{media/}}     
\usepackage{url}
\usepackage{cite}
\usepackage{amsmath}
\usepackage{tikz}
\usepackage{caption}
\usepackage{amsfonts}
\usepackage{amssymb}
\usepackage{algorithm}
\usepackage{longtable}
\usepackage{algorithmicx}
\usepackage{algpseudocode}
\usepackage{multirow}
\usepackage{booktabs} 
\usepackage{diagbox}
\usepackage{balance}
\usepackage{graphicx}
\title{A Mallows-like  Criterion for Anomaly Detection with  Random Forest Implementation
}

\author{
  Gaoxiang Zhao \\
  Department of Mathematics \\
  Harbin Institue of Technology(Weihai) \\
  \texttt{75477845@qq.com} \\
   \And
  Lu Wang \\
  Department of Mathematics and Statistics \\
  Shandong University \\
  \texttt{997546958@qq.com} \\
   \And
  Xiaoqiang Wang\textsuperscript{*} \\
  Department of Mathematics and Statistics \\
  Shandong University \\
  \texttt{xiaoqiang.wang@sdu.edu.cn} \\
}

\begin{document}
\maketitle

\begin{abstract}
The effectiveness of anomaly signal detection can be significantly undermined by the inherent uncertainty of relying on one specified model.  Under the framework of model average methods, this paper proposes a novel criterion to select the weights on aggregation of multiple models, wherein the focal loss function accounts for the classification of extremely imbalanced data. This  strategy is further integrated into  Random Forest algorithm by replacing the conventional voting method. 
We have evaluated the proposed method on benchmark datasets across various domains, including network intrusion. The findings indicate that our proposed method not only surpasses the model averaging with typical loss functions but also outstrips  common anomaly detection algorithms in terms of  accuracy and robustness.
\end{abstract}


\section{Introduction}
Anomaly Detection (AD)~\cite{HA01},\cite{02},\cite{03} aims at identifying extreme data or anomalies from the observed data that are different from normal observations. Applications of AD include financial analysis~\cite{04}, cybersecurity~\cite{05}, paramedical care~\cite{06}, etc. Among the traditional anomaly detection methods, it is often assumed that anomalies are outliers or low probability~\cite{07} points, whereby anomalies are distinguished by attributes based on statistical properties, distance density, and so on. However, these assumptions probably engender two types of limitations. Firstly, the inherent uncertainty associated with any single model can readily lead to fitting issues. Secondly, the consideration is confined to a singular type of anomaly detection, which can weaken the efficiency of anomaly detection, especially when the true outlier type and the method's assumed type differ significantly.

Ensemble method could account for the different type of data and model, but conventional vote or mean approaches may prove inefficient for the extremely imbalance data. Substantially,  model average methods~\cite{08},\cite{09},\cite{10},\cite{11},\cite{12}, have been validated to improve the overall prediction performance by combining the prediction results of multiple base models. Model average methods can be broadly distinguished into two principal techniques: those based on the minimum loss function~\cite{13},\cite{14} and those based on Bayesian methods~\cite{15}. Some researchers have intensively investigated model averaging methods for high-dimensional regression problems, by removing constraint of weight or handle the problem in the presence of responses missing at random~\cite{16},\cite{17}, while others have investigated the improvement of Bayesian model averaging methods, such as combing Bayesian model with selection of regressors~\cite{18}, to enhance interpretability and efficiency of Bayesian model averaging estimation. Some researchers have also explored the application of model averaging in the field of deep learning, and proposed an efficient protocol for decentralized training of deep neural networks from distributed data sources~\cite{19}, which has achieved good results in several deep learning tasks. In the model averaging method based on minimizing the loss function\cite{13},\cite{14}, the loss function is usually selected by choosing the logarithmic loss function~\cite{20}, the squared loss function\cite{21}, and the cross-entropy loss function\cite{22}. Nevertheless, there are few studies leveraging the model averaging method to address the issue of anomaly detection.

The focal loss function~\cite{23},~\cite{24},~\cite{25} is a type of loss function used to address class imbalance issues, particularly in tasks such as object detection and image segmentation. The focal loss function can assign different weights to samples or classes, but its application in the field of model averaging, specifically in assigning weights to base models, has not been explored. In this letter,  we propose a method for optimizing the weights in model averaging by embedding this focal loss function into the Mallows' criterion. More precisely, in random forests,  we add a complexity penalty term to the focal loss function similar to Mallows averaging method\cite{12}, and then assign weights to the sub-decision trees by minimizing Mallows-like criterion. The advantages of this method are as follows: it enhances the ability to handle extremely imbalanced data by using the focal loss function, thereby improving anomaly detection performance; it suppresses model complexity and enhances generalization by adding regularization term to the focal loss function; and it employs model averaging to combine the effects and performances of different base models and allocate different weights, thereby achieving the construction of a higher-precision anomaly detection model.

The proposed Mallows-like focal loss approach is compared with anomaly detection methods based on minimizing other loss functions as well as commonly methodologies for the detection of anomalies. The AUC metric for assessing binary classification performance, the ARI metric for assessing clustering algorithm performance, and the Recall metric, which is important in anomaly detection methods are adopted to ascertain the performance of our proposed methodology. In the evaluation conducted on the KDDCup network intrusion dataset. The proposed approach shows a $2.1\%$ improvement in AUC in comparison to the suboptimal model averaging method based on minimizing the cross-entropy loss function. In terms of the ARI metric for measuring clustering effectiveness, the proposed approach exhibits a $2.5\%$ improvement over the suboptimal voting method. Regarding Recall metric, the proposed approach demonstrates a $5.3\%$ enhancement compared to the suboptimal model averaging method based on minimizing the cross-entropy loss function. Compared with several common anomaly detection methods, our approach has also demonstrated superior performance. To further validate the efficacy of our algorithm more effectively, public benchmark datasets were also employed for testing, with results indicating improved accuracy and stability in anomaly detection.

The main contributions of this letter can be summarized as follows:

\begin{itemize}
\item
We propose a novel  Mallows-like averaging criterion to optimize the weights on aggregation of multiple models, wherein the focal loss function has been instrumental in enhancing the performance of anomaly detection.
\item
Utilizing the Mallows-like focal loss criterion (MFL), we introduce a variant of the Random Forest algorithm, specifically tailored for anomaly detection, within the framework of model averaging for optimal weight selection.
\end{itemize}

\section{PROPOSED METHOD}
In the context, anomaly detection will be considered as a binary supervised classification problem. We summerize the training sample into a set of  $D=\{(Y_1, \boldsymbol x_1), \cdots, (Y_n, \boldsymbol x_n)\} \subseteq \{0,1\} \times \mathbb R^p$, where $\boldsymbol x_i= (x_{i1}, \cdots, x_{ip})\rq{}$  is the predictors of dimension $p$ and the response variable $Y_i = 1$ that represents that $i$-th sample is abnormal, otherwise zero for $i=1,\cdots, n$. The relationship between all  variables can be formulated as
\begin{equation}
	\label{eq:model}
Y_i = f(\boldsymbol x_i) + \varepsilon_i, \quad i =1, \cdots, n,
\end{equation}
where the function $f(\cdot)$ remain unspecified or even non parametric. We suppose that all the residual errors $\varepsilon_i$ are independents and homogenous with $E(\varepsilon_i)=0,
$ $E(\varepsilon^2_i) = f(\boldsymbol x_i)(1-f(\boldsymbol x_i)).$ 

Notice that in Model~\eqref{eq:model}, the $p$ predictors $\boldsymbol x$ can be randomly selected, thereby generating diverse models.  Model averaging method involves the weighted ensemble of models corresponding to each variable selection, that is 
\begin{equation}
f(\boldsymbol x_i) = \sum_{m=1}^{M} \omega_m f_m (\boldsymbol x_i),
\end{equation}
followed by the minimization of a penalized loss function to optimize the selection of optimal weights within the following unit simplex:
\begin{equation}
\mathcal H = \left\{ \boldsymbol{\omega} = (\omega_1,\cdots, \omega_M) \in [0,1]^M \mid \sum_{m=1}^{M} \omega_m =1\right\}
\end{equation}

In the other hand, the focal loss is initially designed to address the object detection scenario for the extreme imbalance data, which adding a modulating factor to the standard cross entropy criterion as follows
\begin{equation}
\begin{split}
FL(Y, f(\boldsymbol x)) = &-Y\alpha(1-f(\boldsymbol x))^\gamma\log(f(\boldsymbol x)) \\
&- (1-Y)(1-\alpha)f(\boldsymbol x)^\gamma\log(1-f(\boldsymbol x)),
\end{split}
\label{eq:focal}
\end{equation}
where $f(\boldsymbol x)$ represents the estimated probability for the class with label $Y=1.$ Incorporating the focal loss into the ensemble method, we propose a novel Mallows-like criterion to determine the optimal weights. Furthermore, this Mallows-like focal loss criteria is realized in Random Forest algorithm. Figure 1 depicts the overall framework of our proposed anomaly detection method, which is a model averaging structure based on minimizing MFL  criterion.
\begin{figure}
\centerline{\includegraphics[width=\columnwidth]{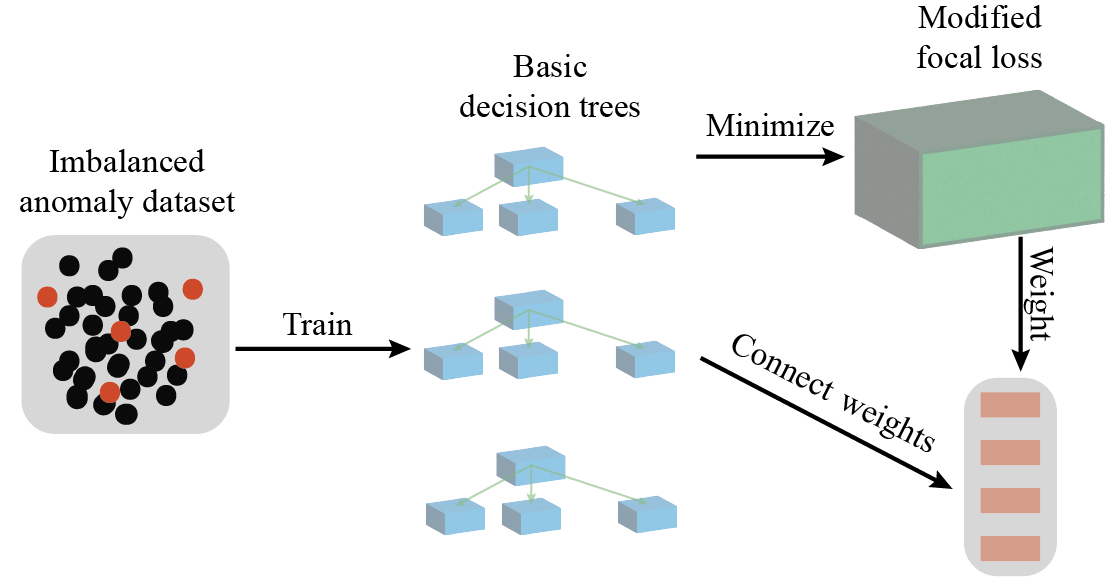}}
\caption{The schematic diagram of the proposed model averaging method  is presented. This approach minimizes MFL criterion to allocate weights to base decision trees, mitigating the effects of data imbalance while controlling model complexity.}
\end{figure}

\subsection{Mallows-like Focal Loss Criterion}
Hansen~\cite{13} (2007) firstly investigated Mallows criterion into the  least squares model averaging  to select the weight vector as 
\begin{equation}
\boldsymbol{\omega}^* = \arg\min_{\boldsymbol{\omega} \in \mathcal H}
\left\{
||Y -  P(\boldsymbol{\omega})Y||^2 + 2\sigma^2 \mathrm{trace}\left(P(\boldsymbol{\omega}) \right)
\right\}\end{equation} \text{with}
$ P(\boldsymbol{\omega}) = \sum_{m=1}^{M}\omega_m P_m,
$
where $\sigma^2$ is an unknown parameter to estimate, and $P_m$ is the projection matrix in linear regression for $m$-th model. Here, the term of $\mathrm{trace}(P(\boldsymbol{\omega})) $ is  defined as the effective number of parameters, that is, weighted average of the number of predictor in each submodel.  
The optimized function consists of two parts: first term measure the fitting error of weighted model; second term penalize the complexity of model.

By substituting the first term in Mallows criterion with focal loss~\eqref{eq:focal}, we propose a special criterion dedicated to anomaly detection as follows 
\begin{equation}
C(\boldsymbol{\omega})= \sum_{i=1}^{n}FL(Y_i, f(\boldsymbol{x}_i)) + 2 \sigma^2(\boldsymbol\omega)\left(1+\sum_{m=1}^{M}\omega_m k_m\right),
\label{eq:criteria}
\end{equation}
where $k_m$ is the number of predictors in $m$-th base model. Moreover, depending on the implemented algorithm, $ \sum_{m=1}^{M}\omega_m k_m$ can be relaxed to a function of the number of predictors in each base model.

Consider the methods of Random Forests and Least Squares Support Vector Classification: in the context of Random Forests, $k_m$ denotes the number of internal nodes within the $m$-th trained decision tree, whereas in Least Squares Support Vector Classification, it represents a measure of the magnitude of support vector weights. More specifically, for the $m$-th classifier, $k_m$ takes the following form:

\begin{equation}
\text{trace}\left(\left(HH^{\top}+\lambda I\right)^{-1}HH^{\top}\right)
\end{equation}

The matrix \( H \) represents the two-dimensional mapping of support vectors relative to the entire sample set via the kernel trick, while \( \lambda \) is a selected parameter indicating the strength of regularization.

Note that the unknown parameter $\sigma^2(\boldsymbol\omega)$ in ~\eqref{eq:criteria} represents the variance of model in Mallow's criterion, we replace  this term by $\hat\sigma^2(\boldsymbol\omega) = \sum_{i=1}^{n} FL(Y_i, f(\boldsymbol{x}_i))/n$ in practice. 

\subsection{Random Forest with MFL}
Random Forest has become a very popular classifiction method due to its great flexibility and promising accuracy. In Random Forest, the voting mechanism is frequently utilized for the classification of data. In contrast, the first term in MFL criterion~\eqref{eq:criteria} implemented in Random Forest measures the fitting error of the weighted random forest in the training sample. The second term in criterion~\eqref{eq:criteria} penalizes the complexity of trees in the forest, says $\sum_{m=1}^{M}\omega_m k_m$ represents the weighted number of leaf nodes of all trees.

Utilizing the MFL criterion~\eqref{eq:criteria}, we propose to realize this algorithm by two following steps:  Initially, we establish $M$ decision trees, denoted as  $\hat f_m(\boldsymbol{x}), m=1,\cdots, M$. Subsequently, we apply the aforementioned criterion to  opimize the weight, represented  by $\boldsymbol{\omega}^*$.  The weighted base models are linearly combined to form the overall model.  Algorithm~\ref{alg:algorithm} illustrates  the execution steps and logical structure of the model averaging method. 
\begin{algorithm}[htbp]
	\caption{Random Forest with Mallows-like focal loss criterion}
	\label{alg:algorithm}
	\begin{algorithmic}
		\Require Dataset $D=\left\{(\boldsymbol x_i, Y_i), i=1,\cdots, n\right\}$, number of subtrees $M$, hyperparameters $\alpha, \gamma.$
		\Ensure Optimal weights $\boldsymbol w^*$
		\While{decision tree $m = 1, \cdots, M$}
		\State 	Create a bootstrap sample $D_{m}$ of size $n$ from the training data $D$
		\State Randomly select  $[\sqrt{M}]$ predictors
		\State Build a decision tree $\hat f_m$ with $k_m$ leaf nodes
		\EndWhile
		
		\State Optimize the weights $\omega$
		{
\small
$$
\boldsymbol\omega^* = \min_{\omega \in \mathcal H}\left\{\sum_{i=1}^{n} FL(Y_i,  f(\boldsymbol{x}_i)) +2 \hat\sigma^2(\boldsymbol\omega)\left(1+\sum_{m=1}^{M}\omega_m k_m\right) \right\}
$$
}
\text{with}
\begin{align*}
& f(\boldsymbol{x}_i) = \sum_{m=1}^{M}\omega_m\hat f_m(\boldsymbol{x}_i), \\
& \hat\sigma^2(\boldsymbol\omega) = \sum_{i=1}^{n} FL(Y_i, f(\boldsymbol{x}_i))/n.
\end{align*}
 \end{algorithmic}
\end{algorithm}

Note that there are two  hyperparameters $\alpha, \gamma$ in focal loss function~\eqref{eq:focal} and the number of subtrees $M$ in random forest algorithm. We adopt Bayesian hyperparameter estimation methods~\cite{26} to expedite training speed and optimize training results. As the focal loss function represents a nonlinear constrained optimization problem, this paper employs sequential least squares methods~\cite{27} to optimize the modified focal loss function, thereby controlling complexity and computational costs. 

\section{EXPERIMENTS}
\label{sec:guidelines}
Anomaly detection typically addresses the issue of severe imbalance between positive and negative samples, for which there is often no clear definition about the proportion. To accelerate training process and increase the ratio of positive to negative samples for validating the model's effectiveness in anomaly detection, we employ hierarchical sampling methods~\cite{28} on the imbalanced dataset, controlling the proportion of minority samples within $5\%$.
We first consider the KDDCup network intrusion dataset~\cite{29}, which is a publicly available dataset in the field of anomaly detection, containing $125,870$ records with $41$ features each. We proportionally extracted $1,069$ records of which the anomalies account for $2.9\%$. Bayesian hyperparameter optimization was employed with focal loss and initial weights set to be $(\alpha, \gamma) = c(0.95, 2)$ and $\omega_m=0.05, m=1,\cdots, M$, respectively. To validate the effectiveness of the proposed method on Random Forest, we compared it with ensemble methods such as voting, model averaging based on minimizing cross-entropy loss, and commonly used methods in anomaly detection such as isolation forest and logistic regression. The dataset was divided into a training set ($70\%$) and a test set ($30\%$), and model was trained $60$ times. The proposed method was evaluated using metrics such as AUC, ARI, accuracy, and recall.
\begin{figure}[htbp]
    \centering
    \includegraphics[width=\columnwidth]{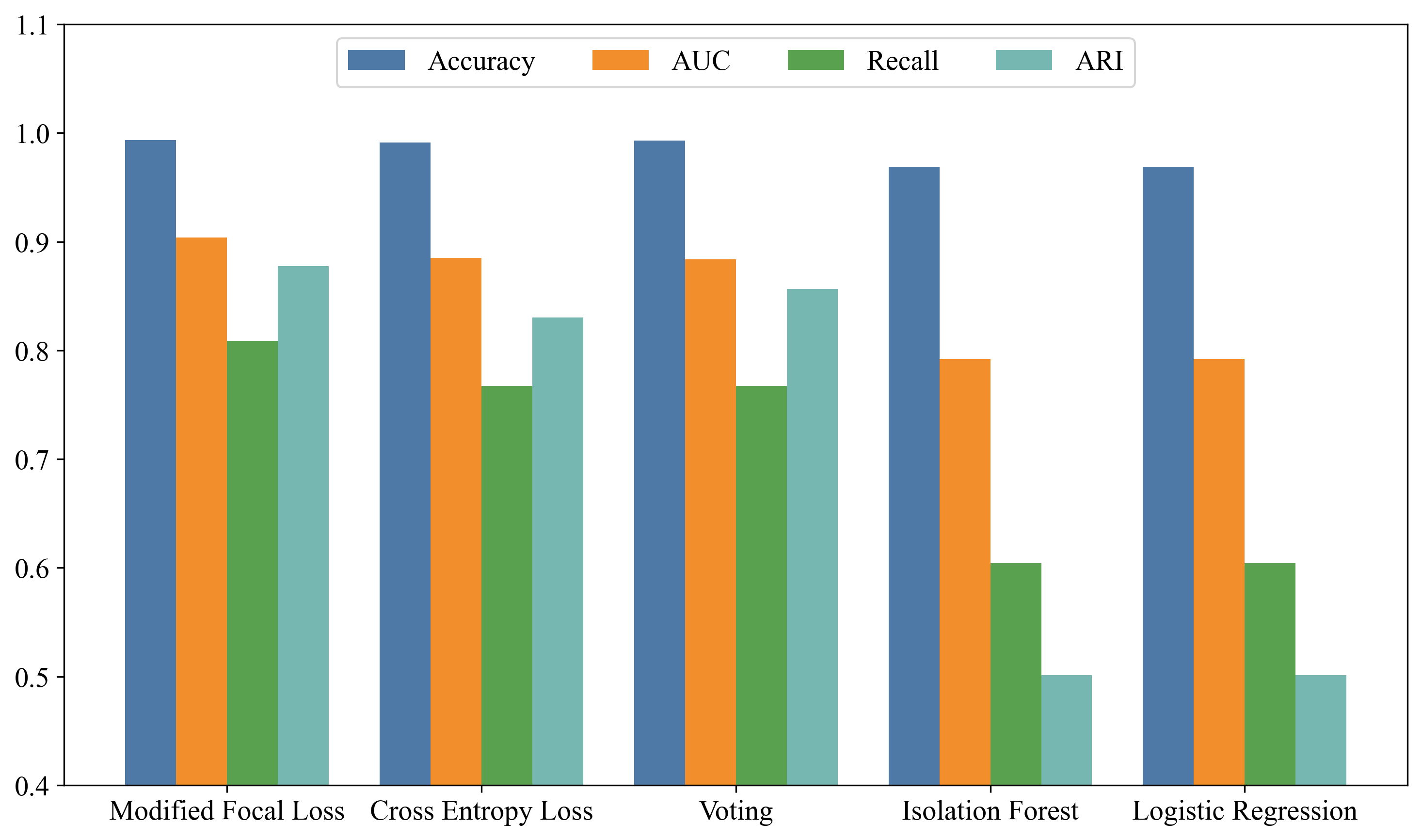}
    \captionsetup{justification=centering}
    \caption{Network intrusion dataset methodology metrics.}
    \label{fig:network}
\end{figure}

Figure~\ref{fig:network} illustrates the performance of the model averaging anomaly detection method based on minimizing MFL criterion on the test set, with average accuracy, AUC, recall, and ARI values of $0.9939,$ $0.9040,$ $ 0.8083,$ and $0.8774$, respectively. On the network intrusion dataset, our proposed method achieved the optimal performance in all evaluation metrics, with improvements of $0.10\%,$ $2.14\%,$ $ 5.33\%$ and $2.45\%$ compared to the second-best metrics, respectively.

To further validate the method's effectiveness for anomaly detection and extremely imbalanced data, we selected ten imbalanced datasets from databases such as UCI~\cite{30}. These datasets span various domains including medicine, industrial production, agricultural production, and image classification. Hierarchical sampling method was applied to these datasets to keep the proportion of minority samples under $5\%$. Subsequently, the datasets were split into a training set ($70\%$) and a test set ($30\%$). After adjusting hyperparameters using Bayesian hyperparameter tuning, the models were trained for 60 rounds.
The proposed method was evaluated using metrics AUC and ARI. 

\begin{table*}[htbp]
\centering
\caption*{TABLE \MakeUppercase{\romannumeral1} \par \textsc{AUC score comparison of anomaly detection algorithms in ten benchmark datasets}}
\label{tab:AUC}
\begin{tabular}{l|cccccccccc}
\toprule
\multirow{2}{*}{\diagbox{Model}{Dataset}} & SB & Pis & MHR & PS & PCO & Ye & Cl & Ca & MF & Sa \\
& & & & & & & & & & \\
\midrule 
Modified Focal & \bfseries{0.7847} & \bfseries{0.6652} & \bfseries{0.7717} & \bfseries{0.7408} & \bfseries{0.8598} & \bfseries{0.6214} & \bfseries{0.5666} & 0.6106 & \bfseries{0.9333} & \bfseries{0.9794} \\
Focal & 0.7843 & 0.6641 & 0.7703 & 0.6997 & 0.8287 & 0.5993 & 0.5655 & 0.6107 & 0.9317 & 0.9458 \\
Vote & 0.6492 & 0.6298 & 0.7455 & 0.6736 & 0.8156 & 0.5426 & 0.5334 & 0.5757 & 0.9072 & 0.9377 \\
Zero One & 0.6392 & 0.619 & 0.7395 & 0.6679 & 0.8124 & 0.545 & 0.5315 & 0.5698 & 0.905 & 0.9387 \\
Hamming & 0.6394 & 0.619 & 0.7395 & 0.6679 & 0.8124 & 0.545 & 0.5315 & 0.5698 & 0.905 & 0.9387 \\
Hinge Loss & 0.695 & 0.6546 & 0.7485 & 0.6939 & 0.806 & 0.5832 & 0.563 & \bfseries{0.6113} & 0.9129 & 0.9297 \\
Cross Entropy & 0.6885 & 0.6495 & 0.7619 & 0.6842 & 0.8113 & 0.5778 & 0.5603 & 0.6077 & 0.9169 & 0.9366 \\
Average & 0.6327 & 0.6135 & 0.7335 & 0.6597 & 0.812 & 0.5401 & 0.5296 & 0.5671 & 0.9017 & 0.9371 \\
Isolation Forest & 0.6251 & 0.5038 & 0.6421 & 0.6595 & 0.621 & 0.5396 & 0.5137 & 0.5448 & 0.4767 & 0.4777 \\
logistic & 0.6659 & 0.5 & 0.5 & 0.5 & 0.5 & 0.5 & 0.5 & 0.5 & 0.5 & 0.5 \\
KNN & 0.5292 & 0.5971 & 0.6498 & 0.6128 & 0.6846 & 0.5141 & 0.5302 & 0.5097 & 0.836 & 0.9555 \\
\bottomrule 
\end{tabular}
\end{table*}

\begin{table*}[htbp]
\centering
\caption*{TABLE \MakeUppercase{\romannumeral2} \par \textsc{ARI score comparison of anomaly detection algorithms in ten benchmark datasets}}
\label{tab:ARI}
\begin{tabular}{l|cccccccccc}
\toprule
\multirow{2}{*}{\diagbox{Model}{Dataset}} & SB & Pis & MHR & PS & PCO & Ye & Cl & Ca & MF & Sa \\
& & & & & & & & & & \\
\midrule
Modified Focal & \bfseries{0.6246} & \bfseries{0.3881} & \bfseries{0.6106} & \bfseries{0.469} & 0.6465 & \bfseries{0.2738} & \bfseries{0.1668} & 0.2816 & \bfseries{0.9211} & 0.8616 \\
Focal & 0.6239 & 0.3863 & 0.6099 & 0.429 & 0.6555 & 0.2532 & 0.1649 & \bfseries{0.2823} & 0.9191 & 0.8954 \\
Vote & 0.4254 & 0.3476 & 0.5894 & 0.4161 & \bfseries{0.69} & 0.1345 & 0.0938 & 0.2292 & 0.8873 & 0.9009 \\
Zero One & 0.4016 & 0.327 & 0.5834 & 0.4093 & 0.6852 & 0.1416 & 0.0887 & 0.2138 & 0.8846 & 0.9011 \\
Hamming & 0.4021 & 0.327 & 0.5834 & 0.409 & 0.6852 & 0.1416 & 0.0887 & 0.2138 & 0.8846 & 0.9011 \\
Hinge Loss & 0.4755 & 0.336 & 0.5634 & 0.4122 & 0.6322 & 0.2067 & 0.1504 & 0.279 & 0.8588 & 0.8503 \\
Cross Entropy & 0.4961 & 0.3641 & 0.5869 & 0.4105 & 0.6615 & 0.2202 & 0.154 & 0.2677 & 0.8698 & 0.8849 \\
Average & 0.386 & 0.3148 & 0.5754 & 0.3965 & 0.6867 & 0.1285 & 0.0836 & 0.2059 & 0.8799 & 0.9 \\
Isolation Forest & 0.2171 & 0.0056 & 0.2416 & 0.2387 & 0.2687 & 0.065 & 0.0164 & 0.0667 & -0.0308 & -0.037 \\
logistic & 0.4199 & 0 & 0 & 0 & 0 & 0 & 0 & 0 & 0 & 0 \\
KNN & 0.0953 & 0.2669 & 0.4036 & 0.3125 & 0.4559 & 0.0324 & 0.0859 & 0.0277 & 0.7838 & \bfseries{0.9293} \\
\bottomrule
\end{tabular}
\end{table*}

Table \MakeUppercase{\romannumeral1} compares the AUC values of the proposed MFL method with other anomaly detection algorithms. The results indicate that the random forest employing the MFL criterion exhibits superiority in AUC evaluation across nine datasets. Compared to the focal loss based model averaging method without penalty, MFL method shows significant advantages on four datasets. These findings validate the effectiveness of our method.

Table \MakeUppercase{\romannumeral2}  compares the ARI metrics of the MFL method with other anomaly detection algorithms across the datasets. Notably, the MFL method demonstrates superior performance on seven datasets. KNN method, voting ensemble method, and model averaging method based on minimizing focal loss show optimal performance on the rest three datasets respectively. These findings highlight the significant advantage of the MFL method.

\section{Discussion}
In this letter we propose a Mallows-like model averaging criterion for anomaly detection based on the focal loss function. This Mallows-like  focal loss criterion is later implemented in Random Forests algorithm to address data extremely imbalance issues.
To validate the performance of our proposed method, we selected  public benchmark datasets to compare the proposed approach against the other ensemble models, as well as other commonly used anomaly detection methods. All the results indicate that our method has exhibited superior performance in terms of both anomaly recall and classification accuracy.

In the future, Mallows-like focal loss criterion in heteroscedasticity could be investigated, where the variance term in Mallows criterion may be replaced by the residuals for each sample.  
Moreover, the performance of the proposed approach is constrained by the selection of hyperparameters in focal loss. A possible future research direction involves designing efficient hyperparameter tuning methods tailored for anomaly detection algorithms and providing theoretical support.

\bibliographystyle{unsrt}  
\bibliography{references}

\end{document}